\documentclass[runningheads]{llncs}
\usepackage{graphicx}
\usepackage{amsmath}
\usepackage{array}
\usepackage[misc]{ifsym}
\begin{document}
%
\title{Semi-interactive Attention Network for Answer Understanding in Reverse-QA}
%
%

\author{Qing Yin\inst{1} \and
Guan Luo\inst{2} \and
Xiaodong Zhu\inst{3} \and
Qinghua Hu\inst{1} \and
Ou Wu\textsuperscript{1(\Letter)}}
\authorrunning{Q. Yin et al.}
%
\institute{Tianjin University, Tianjin 300110, China \\
\email{\{qingyin, huqinghua, wuou\}@tju.edu.cn}\and
NLPR, Chinese Academy of Sciences, China.
\email{gluo@nlpr.ia.ac.cn}\and
University of Shanghai for Science, China.
\email{zhuxd81@gmail.com}}

\maketitle

\begin{abstract}
Question answering (QA) is an important natural language processing (NLP) task and has received much attention in academic research and industry communities. Existing QA studies assume that questions are raised by humans and answers are generated by machines. Nevertheless, in many real applications, machines are also required to determine human needs or perceive human states. In such scenarios, machines may proactively raise questions and humans supply answers. Subsequently, machines should attempt to understand the true meaning of these answers. This new QA approach is called reverse-QA (rQA) throughout this paper. In this work, the human answer understanding problem is investigated and solved by classifying the answers into predefined answer-label categories (e.g., $True$, $False$, $Uncertain$). To explore the relationships between questions and answers, we use the interactive attention network (IAN) model and propose an improved structure called semi-interactive attention network (Semi-IAN). Two Chinese data sets for rQA are compiled. We evaluate several conventional text classification models for comparison, and experimental results indicate the promising performance of our proposed models.

\keywords{Question answering\and Reverse-QA\and Attention\and LSTM.}
\end{abstract}
\section{Introduction}
Question answering (QA) is applied in many real applications, such as robots and intelligent customer service. The goal of QA is to provide a satisfactory answer depending on users' question \cite{Kumar}. QA can provide a more natural way for humans to acquire information than traditional search engines \cite{Hixon}.

In nearly all existing QA studies and applications, the questions are raised by humans and the answers are generated by machines. In other words, in existing QA, humans are the questioners and machines are the answerers. Therefore, selecting a satisfactory answer from candidate answer corpora, which are also known as answer selection, is the key problem in QA \cite{Tan}. In addition to meeting users' information requirements, machines in some real applications, such as telephone survey \cite{Lipps}, are also required to actively acquire the exact needs or feedbacks of users. Accordingly, machines may choose to proactively raise questions to users and then analyze their answers. In other words, machines are the questioners and humans are the answerers. This process is a reverse of the conventional QA process and is called reverse-QA (rQA) in this paper. Fig. 1 shows the conventional QA and rQA processes.

In conventional QA, the key problem is understanding users' questions. On the contrary, the key problem in rQA is understanding users' answers. In the present study, two types of machine-launched questions are considered, namely, true-or-false (T/F) and multiple-choice (MC) questions. Table 1 shows two illustrative examples for T/F and MC questions. Some human answers in Table 1 are easy to analyze. For example, the ``Yes" and ``No" answers to the first question can clearly distinguish the category. However, other answers are vague and difficult to process for their exact meanings. For example, the answer ``I was a teacher last year" for the second question is a `false' response, but it is easily classified as a `true' response. Hence, understanding users' answers is not a trivial task.
\begin{figure}[ht]
\centering
\includegraphics[scale=0.6]{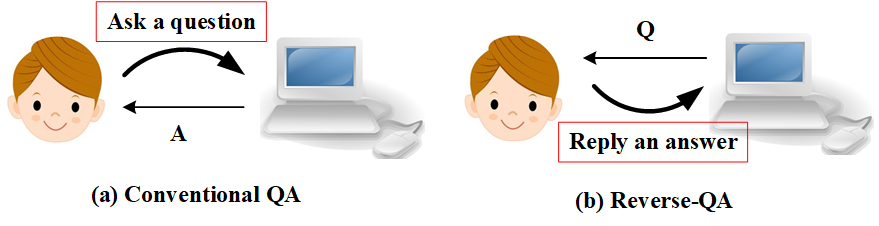}\vspace{-0.16in}
\caption{Difference between conventional QA (a) and reverse-QA (b).}
\label{fig:label}
\end{figure}\vspace{-0.21in}
\begin{table}
\caption{Illustrative examples of the T/F and the MC questions.}\label{tab1}\vspace{-0.11in}
\centering
\begin{tabular}{|l|p{5cm}|p{6cm}|}
\hline
Type &  \centering{Question} & (Possible) answers by human\\
\hline\hline
T/F &  {Do you like running?} & Yes/A little/No/Sometimes.\\
T/F&  { Are you a teacher?} & Sometimes/I'm not sure/You guess/ I was a teacher last year.\\\hline
MC & { Would you like coffee or tea?} & Coffee/Tea/No, thanks/Either is ok.\\
MC & { Are you usually walking, cycling, or driving to work?} & Walk/Except cycling/I lost my job/By train/It all depends.\\
\hline
\end{tabular}\vspace{-0.11in}
\end{table}

\renewcommand{\thefootnote}{\fnsymbol{footnote}}
As far as we know, it is the first to focus on the rQA procedure and corresponding answer understanding. Considering that no public data set is available for this work, two data sets\footnote{These two data sets have been uploaded to Github and the URL is provided after anonymous review.} are compiled to construct and test the models for rQA. We simply take the answer understanding in rQA as a classification task and use several common text classification techniques. These classification algorithms ignore the relationships between a (machine) question and an (human) answer. To this end, two new models based on deep neural network (DNN) are proposed. The first model is based on the interactive attention network (IAN) \cite{Ma}. The second model is a simplified but a more effective version of IAN and is called semi-interactive attention network (Semi-IAN). The experimental results suggest the potential of the proposed models. The contributions of this work are summarized as follows:
\renewcommand{\labelitemi}{$\bullet$}
\begin{itemize}
\item We investigate a new QA procedure called rQA. Moreover, a new problem called answer understanding for rQA is proposed. Two data sets are collected and labeled depending on two common question types, namely, T/F and MC. The two data sets can be used to construct and evaluate new models.

\item Two new models are proposed to capture the semantic relationships between questions and answers. The proposed IAN model is based on the raw IAN \cite{Ma}, which is initially designed for opinion mining. The proposed Semi-IAN model, which considers questions as background, achieves highest accuracy throughout the experiments.
\end{itemize}
\section{Related Work}
\subsection{QA}
QA is a crucial NLP task that depends on natural language understanding and domain knowledge \cite{Xiong}. Given a question from users, QA returns an answer via answer selection or generation based on a knowledge base. In most existing QA studies, the answer selection is implemented by the matching between a question and candidate answers or documents. The answer that has the highest match score is usually selected and returned to users. According to the matching procedure, most existing QA methods can be divided into two categories:

\begin{itemize}
  \item Hard-crafted feature-based methods. This category of methods extracts lexical features to represent questions and candidate answers \cite{Richardson}\cite{Wang700}. Chen et al.  \cite{Chen} proposed a feature fusion strategy for various features, including carefully crafted lexical, syntactic, and word order features.
  \item Deep feature-based methods. This category of methods extracts deep features via a CNN or long-short time memory network (LSTM) \cite{Hill}. Kadlec et al. \cite{Kadlec} presented a pointer-style attention mechanism  for text feature representation in QA.
\end{itemize}
Some other works have focused on questions \cite{Bao} and visual QA \cite{Malinowski} .

\subsection{Text Classification}
Text classification aims to predict the category of an input text sample. The category can be semantic (e.g., political and economic) or sentimental (e.g., positive and negative) \cite{Zhang25}. Besides some rule-based methods \cite{Sasaki}, most existing methods are based on machine learning theories. Nearly all classical (shallow) classifiers have been used in text classification, such as support vector machine (SVM) \cite{Kiritchenko}, KNN \cite{Deng}, and logistic regression (LR), etc.

In recent years, the emergence of deep learning as a powerful technique for nearly all NLP tasks has facilitated the adoption of classical DNNs (e.g., CNN \cite{Zhang} and Recurrent Neural Network (RNN) \cite{Tang}) for these tasks.

\section{Methodology}
Answer understanding in rQA can be formulated into a text classification problem as follows. By considering a machine-question and human-answer pair ($q$, $s$) and a predefined answer-label $A$, we aim to predict the category $c$ ($c$ $\epsilon$ $A$) for $s$.

In this study, two common types of questions are considered, namely, T/F and MC. The two types of questions correspond to two scenarios, namely, T/F and MC rQA. The primary difference between T/F and MC rQA lies in the definitions of the answer-label set $A$.

In the T/F rQA, the answer-label set $A$ can be set as  \{$¡®True¡¯$, $¡®False¡¯$, $¡®Uncertain¡¯$\} regardless of the question. In the MC rQA, we assume that the option set is $I$, and the answer-label set $A$ is the union of all the subsets of $I$ plus the $¡®Uncertain¡¯$ element. For example, if $I$ is \{$¡®opt1¡¯$, $¡®opt2¡¯$, $¡®opt3¡¯$\}, then the set $A$ is \{\{$opt1$\}, \{$opt2$\}, \{$opt3$\}, \{$opt1$, $opt2$\}, \{$opt2$, $opt3$\}, \{$opt1$, $opt3$\}, \{$opt1$, $opt2$, $opt3$\}, $¡®Null¡¯$, $¡®Uncertain¡¯$\}.

The following part introduces how the answer category $c$ is inferred in the two rQA scenarios above.

\subsection{Answer Understanding for T/F rQA}
In T/F rQA, we aim to classify the human answer $s$ into one element (category) of the predefined answer-label set \{$¡®True¡¯$, $¡®False¡¯$, $¡®Uncertain¡¯$\}. Intuitively, most existing classification methods can be used.

\subsubsection{3.1.1 Text Classification-based Methods}
If $s$ or the simple concatenation of $q$ and $s$ is taken as a piece of input texts, then three typical methods are obtained and listed below:
\renewcommand{\labelitemi}{$\bullet$}
\begin{itemize}
\item \textbf{Rule-based method}: This method relies on some key words, such as `$ok$', `$yes$', `$not$'. These key words directly indicate a `true' or `false' answer.
\item \textbf{Bag-of-words}: This method firstly extracts a bag-of-word (BOW) feature vector for the input text and then classifies the texts using conventional shallow classifiers such as SVM \cite{Kiritchenko}, LR.
\item  \textbf{DNN-based method}: This method first extracts a deep feature vector for the input text and then classifies it on the basis of the softmax layer of the involved DNN.
\end{itemize}
\subsubsection{3.1.2 The Proposed Models}
The above-mentioned text classification-based methods independently extract the feature vectors of the question and the answer or extract one feature vector from the texts by simply concatenating the question and the answer. These two strategies simply ignore the semantic relationship between the question and the answer. Intuitively, the question and answer texts can facilitate the analysis of their counterpart. Alternatively, their feature extraction procedures should not be independent.

In opinion mining, Ma et al. \cite{Ma} proposed an IAN to extract features for target and contextual texts. In IAN, the target information is used in feature extraction for contextual texts, and the latter is also used in feature extraction for the former. This network is used with a slight modification to utilize the relationship between a pair of question and answer. The overall architecture is shown in Fig. 2(a).
\begin{figure}[t]
\centering
\includegraphics[scale=0.3]{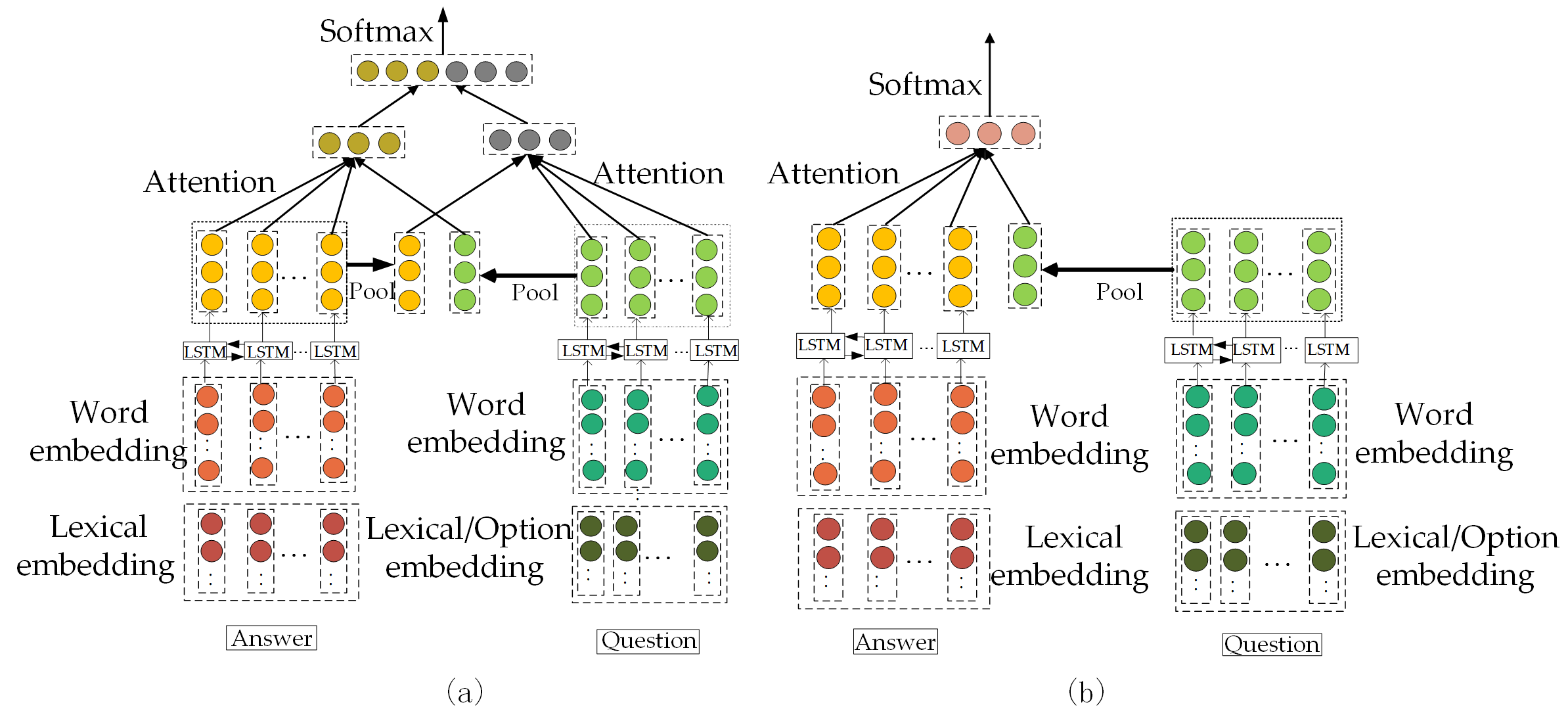}\vspace{-0.08in}
\caption{ (a) IAN for answer understanding in rQA. (b) Structure of the Semi-IAN. (In question modeling part of MC, the additional embedding is option embedding.)}\vspace{-0.11in}
\label{fig:labe2}\vspace{-0.08in}
\end{figure}

We let $w_{q}^t$ and $w_{s}^t$ denote the $t$th words in the question $q$ and the answer $s$, respectively. The embeddings of $w_{q}^t$ and $w_{s}^t$ consist of two parts. The first part is word embedding, and the second part is lexical embedding. The word embedding is implemented by the standard word2vec algorithm. The lexical embedding relies on a $\rho$-hot encoding \cite{Wu} on a pre-compiled dictionary of several key words. In this study, our dictionary contains six classes of key words: affirmative, privative, suspicious, positive, negative, supposed.

The two input embedding sequences are then fed into a bi-directional LSTM to infer the hidden representation of each word in a sentence. In our model, the forward LSTM at the $t$th input word is as follows:
\begin{equation}
\begin{split}
&\ i_t = \sigma ({W_i}[{c_{t - 1}},{h_{t - 1}},{x_t},{l_t}] + {b_i})\\
&\ f_t = \sigma ({W_f}[{c_{t - 1}},{h_{t - 1}},{x_t},{l_t}] + {b_f})\\
&\ o_t = \sigma ({W_o}[{c_t},{h_{t - 1}},{x_t},{l_t}] + {b_o})\\
&\ d_t = \sigma ({W_d}[{c_{t - 1}},{h_{t - 1}},{x_t},{l_t}] + {b_d})\\
&\ {c_t} = {i_t} \otimes {d_t} + {f_t} \otimes {d_{t - 1}}\\
&\ {h_t} = {o_t} \otimes \tanh ({c_t})
\end{split}
\end{equation}
where $x_t$ and $l_t$ are the word and lexical embedding for the $t$th word, respectively; $i_t$ and $d_t$ are the input vectors of the input unit and the input gate, respectively, for the $t$th word; $o_t$ and $h_t$ are the output and hidden vectors, respectively; $f_t$ is the output of the forget gate; $c_t$ is the internal state of the memory cell in a LSTM unit; $\sigma$ is the sigmoid active function. The backward LSTM is very similar to the forward one except that the input sequence is fed in a reversed way. The output of the bi-LSTM and the lexical embedding of words are concentrated.

After the hidden vectors for each input word are obtained, two pooling vectors $q_{avg}$ and $s_{avg}$ are produced for $q$ and $s$, respectively. These two pooling vectors are used to calculate the interactive attention weights. Let $h$ = [$h_{s}^1$, $h_{s}^2$, $\cdots$, $h_{s}^n$] be the hidden vectors for the answer $s$. Following the definition in \cite{Ma}, the attention weight for the $t$th hidden vector $h_s^t$ is calculated as follows:
\begin{equation}
{\alpha_t} = \frac{{exp(\gamma (h_s^t,{q_{avg}}))}}{{\sum\nolimits_k {exp(\gamma (h_s^k,{q_{avg}}))} }}
\end{equation}
where $\gamma$ is a score function and defined as follows:
\begin{equation}
\gamma (h_s^t,{q_{avg}}) = tanh(h_s^t \cdot {W} \cdot q_{avg}^T + {b})
\end{equation}
where $W$ and $b$ are the parameters to be learnt. Similarly, Eq. (2) indicates the attention weight for the $t$th hidden vector in the question modeling part when $h_{s}^t$ is changed to $h_{q}^t$ and $q_{avg}$ is changed to $s_{avg}$.

Once the attention weights for the hidden vectors for $q$ and $s$ are obtained, the weighted representations recorded as $q_r$ and $s_r$ can be subsequently obtained. The final feature vector $q_s$ is the concatenation of $q_r$ and $s_r$, i.e., $q_s$ = $[q_r^T, s_r^T]^T$.

The proposed model is based the original IAN proposed by Ma et al. \cite{Ma} with slight modifications. Intuitively, the answer is the focus and the question is the background. Nevertheless, the answer and the question are symmetric and equal in the model shown in Fig. 2. To this end, an improved model is proposed and shown in Fig. 2 (b). Because this model only used partial interactive information, it is called semi-interactive attention network (Semi-IAN).

In the Fig. 2 (b) model, the final feature vector $q_s$ does not contain the feature representation $q_r$ of the question part. Alternatively, the question is only used as a background text in answer understanding. The experimental results show that this Semi-IAN model outperforms the raw IAN.

\subsection{Answer Understanding for MC rQA}
In MC rQA, the size of the answer-label set $A$ depends on the number of candidate options for selection. In other words, the number of categories varies depending on the concrete question. Conventional classification technique is inappropriate for the scenario with varied number of categories. Consequently, the original answer classification in MC rQA should be transformed into a new classification problem with a fixed number of categories. In this work, the transformation is implemented as follows. Without loss of generality, let the option set $I$ of one MC question be \{$option1$, $option2$, $option3$\}. The raw answer classification is transformed into three new classification subtasks. The first sub-task is about $option1$; the second is about $option2$; the third is about $option3$. Each subtask infers an answer category from the set {\emph{True}, \emph{False}, \emph{Uncertain}} for the corresponding option. With the above transformation, the new classification problem is with a fixed number (i.e., three) categories.


The model for the transformed MC question is same as the model introduced in Section 3.1.2 with only one difference that lexical embedding in question modeling part is replaced with option embedding. The option embedding is also based on the $\rho$-hot encoding to indicate the current option to be considered, as shown in Fig. 2. Therefore, if $k$ options exist, the model should be run $k$ times to infer the category for each option depending on the input question and human answers. For example, if the predicted categories for the three options are $¡®Ture¡¯$, $¡®False¡¯$, $¡®True¡¯$, then the final output category $c$ is \{$option1$, $option3$\}.

\section{Experimental Data Construction}
Existing QA and text classification benchmark data sets are inappropriate for training and evaluating rQA models. Therefore, two data sets are compiled with a standard labeling process. The two data sets are named rQAData1 and rQAdata2 for the T/F rQA and MC rQA, respectively. The type of MC questions we studied is limited in the type that the options appear in the question, which we call option-contained MC questions.

For the two data sets, the questions are constructed as follows. First, seven domains are selected, namely, encyclopedia, insurance, personal, purchases, leisure interests, medical health, and exercise. Ten graduate students, specifically six males and four females, were invited to participate in the data compiling using Email advertising from our experimental laboratory. All the participants are Chinese and in the age of [22, 30]. Considering that the question and answer generations are not difficult to understand, we did not give special instructions to the participants. Each participant was allowed to construct 50 to 60 questions. We obtain 20 insurance questions, 30 encyclopedia questions, and 40 questions in each of the remaining areas for T/F rQA; for MC rQA, we obtain 20 questions in the insurance field and 40 questions in each of the remaining categories. Finally, 503 questions are obtained after deleting some invalid questions. Among that, the numbers of questions in rQAData1 ang rQAData2 are 250 and 253, respectively.

The answers are constructed as follows. The 503 questions are equally assigned to the 10 participants, and each question is given 18 to 22 answers. The participants also labeled their answers.

For the rQAData1, the types of answers are roughly divided into affirmative, negative, uncertain, and unrelated. Given that the uncertain and unrelated answers are similar in function to the next question, we classify them as the same class. In this way, each answer is tagged with `1' for true, `0' for false, and `2' for uncertain or unrelated. Each sample consists of three components: question, answer, and label. The numbers of training and testing samples are 4,080 and 1,000, respectively.

For the rQAData2, the number of options for each MC question are different and cannot be categorized uniformly. Thus, we add the option information to the MC questions and get a series of transformed MC questions as described in Section3.2. Therefore, the same answer to the same question will have different labels for dissimilar options. Similarly, `1' indicates that the answer is an `true' answer to the current option, `0' implies that the answer is a `false' answer to the current option, and `2' denotes that the answer is `uncertain' answers to the current option or the answer is meaningless to this question. Each sample consists of four components: question, option, answer, and label. There are 12,923 transformed MC questions, and the numbers of training and testing samples are 9,074 and 3,876, respectively. Table 2 presents the brief summary.

\
\begin{table}
\caption{Statistics of our rQA datasets.}\label{tab2}\vspace{-0.11in}
\centering
\begin{tabular}{|c|c |c|c|}
\hline
Data sets & Train samples & Test samples  & $False$/$True$/$Uncertain$\\
\hline
rQAData1  & ¡®4,047¡¯ & ¡®1,000¡¯ & ¡®2,153/2,266/628¡¯\\
rQAData2 & 9,047 & 3,876 & 6,257/4,872/1,814\\
\hline
\end{tabular}\vspace{-0.11in}
\end{table}

\section{Experiment}
\subsection{Comparative Methods}
To demonstrate the validity of our models, the following methods are considered in the experimental comparison.
\renewcommand{\labelitemi}{$\bullet$}
\begin{itemize}
\item \textbf{Rule-based method}: This method is introduced in Section 3.1.1.

\item \textbf{BOW+}: The main idea of BOW is to extract features with a BOW model and send them to a classifier \cite{Mullen} In the subsequent experiments, we use LR and SVM as the classification algorithms.

\item \textbf{CNN/LSTM/Bi-LSTM (A)}: The CNN/LSTM/Bi-LSTM (A) network is used to extract deep features from answers only. The features are then fed to the softmax classification layer to obtain the possible results of the category.

\item \textbf{CNN/LSTM/Bi-LSTM (A+Q)}: Unlike CNN/LSTM/Bi-LSTM (A), these methods extract deep features from the concatenation of answers and questions.

Our proposed methods are listed as follows:

\item \textbf{IAN$^+$}: This method is introduced in Section 3.1.2. It is similar to the raw IAN model with a slight modification.

\item \textbf{Semi-IAN}: The network structure of this method is shown in Fig. 2(b).
\end{itemize}

\begin{table}
\caption{ Classification accuracies on different training set settings.}\label{tab3}\vspace{-0.11in}
\centering
\begin{tabular}{| c | c | c | c | c | }
\hline
Model & rQAData1(A) & rQAData1(A+Q)  & rQAData2(A) & rQAData2(A+Q)\\
\hline
Ruled-Based  & 0.438 & / & 0.298 & / \\
BOW+LR & 0.486 & 0.473 & 0.318 & 0.314\\
BOW+SVM &0.649 & 0.612 & 0.503 & 0.532\\
CNN & 0.673 & 0.615 & 0.521 & 0.530\\
LSTM & 0.685 & 0.652 & 0.532 & 0.534\\
Bi-LSTM & 0.708 & 0.669 & 0.534 & 0.530\\
IAN$^+$ & / & 0.720 & / & 0.578\\
Semi-IAN & / & \textbf{0.735} & / & \textbf{0.585} \\
\hline
\end{tabular}\vspace{-0.11in}
\end{table}

\subsection{Training Settings}
All the DNN models are trained by applying Keras that is equipped with Tensorflow. Both our data sets use accuracy as the metric. The division of training and test data is shown in Table 2. The specific training settings used in our experiment are listed as follows:
\renewcommand{\labelitemi}{$\bullet$}
\begin{itemize}
\item In BOW, we put words that appear more than twice into the dictionary.

\item For SVM, parameters $C$ and $g$ are searched via five-fold cross validations from \{0.1, 1, 5, 10, 100\} and \{0.01, 0.1, 1, 5, 10\}, respectively. For LR, the codes in MATLAB are used, and all the parameters are set to default.

\item For deep models, the word embedding dimension is set to 300 by GloVe \cite{Pennington}.

\item In both IAN$^+$ and Semi-IAN, the $\rho$-hot encoding \cite{Wu} is used in the lexical and option embeddings. In $\rho$-hot encoding, the size $k$ is searched in [ 1, 2,  4, $\cdots$, 16]; the parameter $\rho$ is searched in [0.1, 0.2, $\cdots$, 1].
\end{itemize}
\subsection{Overall Competing Results}
Table 3 shows the classification accuracies for all competing methods. Among these methods, the rule-based method has the worst effect. The two BOW methods achieve better accuracies than the rule-based method. Although the SVM-based method is 0.2\% higher than some deep network methods, the overall method based on deep learning has a better effect. On both data sets, the accuracies of the LSTM-based method are considerably higher than those of the CNN-based method. This finding illustrates that LSTM is more suitable for the problem investigated in this study than CNN. The underlying reason is that LSTM can effectively extract the semantic expression of text information with complex time correlation and different lengths.

On rQAData1, Bi-LSTM outperforms LSTM over 1\% and 3\%. On rQAData2, the accuracies of Bi-LSTM-based and LSTM-based approaches are same, basically.

In general, the performance of most CNN/LSTM/Bi-LSTM (A+Q) is worse than that of CNN/LSTM/Bi-LSTM (A) on our data sets. The reason is that although the adding of the questions is definitely useful, simple concatenation of answers and questions is not conducive to the judgment of answers. It is necessary to propose new methods of introducing questions.

Compared with the Bi-LSTM (A+Q) model, IAN$^+$ improves the performance by approximately 5.1\% and 4.8\% on the rQAData1 and rQAData2, respectively. We can see that Semi-IAN achieves the best performance. The main reason may be that the IAN$^+$ and Semi-IAN provide a more effective way to combine the answer and question texts than the simple way that discards question texts or directly concatenates answer and question texts. Furthermore, the Semi-IAN model highlights the answer texts compared with IAN$^+$.

In Table 3, the accuracies on rQAData2 are lower than those on rQAData1. The classification problem investigated on rQAData2 contains additional information, that is, the options. Additional information also brings more challenges, so the classification for rQAData2 is more difficult than that for rQAData1. The main reason is the understanding for MC answers requires more domain knowledge. In our feature work, we will introduce knowledge graph for the involved domains into our models.
\subsection{Discussion on the Key Modules in Our Models}
The input embedding and attention modules are crucial for a deep model. The lexical and option embeddings used in our input layer incorporate domain knowledge into the network. We first evaluate this embedding strategy.

Table 4 shows the results of the Bi-LSTM, IAN$^+$ and Semi-IAN with or without additional embedding containing lexical and option embeddings on the two data sets. In the six groups of comparison, four groups showed the effectiveness of additional coding. In particular, there was 1\% improvement in our semi-INA model, indicating that the proposed lexical and option embedding are useful.

\begin{table}
%
\caption{Results of with or without lexical and option embedding.}\label{tab4}\vspace{-0.11in}
\centering
\begin{tabular}{| c | c | c | c | }
\hline
Model & W/O & rQAData1  & rQAData2 \\
\hline
Bi-LSTM (A+Q)  & W & 0.665 & 0.495\\
Bi-LSTM (A+Q)  & O & 0.669 & 0.491\\
IAN$^+$ & W & 0.720 & 0.578\\
IAN$^+$ & O & 0.728 & 0.546\\
Semi-IAN & W & \textbf{0.735} & \textbf{0.585}\\
Semi-IAN & O & 0.726 & 0.575\\

\hline
\end{tabular}
\vspace{-0.11in}
\end{table}
Three important parameters are involved in our lexical and option, namely, size $k$, $\rho$ in lexical embedding, and $\rho$ in option embedding. We record the classification performance with different values of these parameters in the experiments. The green curves in Figs. 3(a) and 3(b) show the accuracy variations in terms of different $\rho$ values in lexical embedding. The blue curve in Fig. 3(b) shows the accuracy variations in terms of different $\rho$ values in option embedding. The three curves indicate that the tuning of the $\rho$ value in lexical embedding is useful. In Figs. 3(c) and 3(d), the accuracies when $k$=14 for rQData1 and $k$=8 for rQData2 are larger than those when $k$=1 for both sets. The tuning for $k$ is also useful.

\begin{figure}[t]
\centering
\includegraphics[scale=0.34]{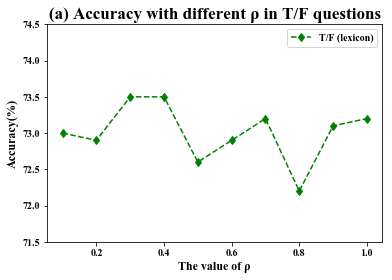}
\includegraphics[scale=0.34]{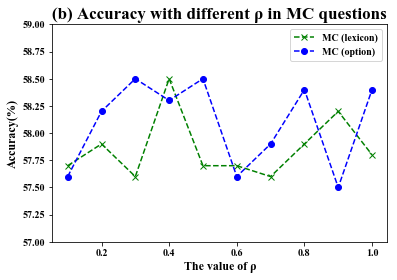}
\includegraphics[scale=0.34]{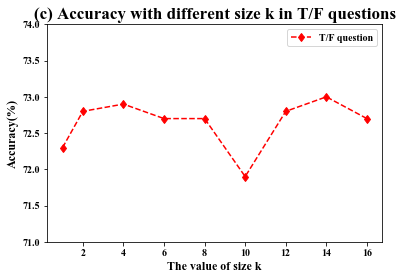}
\includegraphics[scale=0.34]{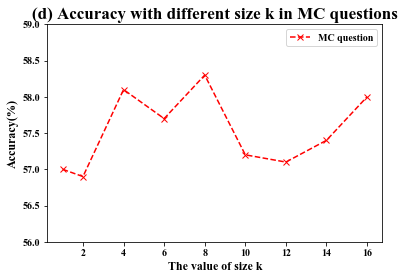}
\vspace{-0.11in}
\caption{Results of accuracy under different parameter values.}
\label{fig:labe3}
\end{figure}

Next, we investigate the effectiveness of a new attention mechanism, namely, CRF attention \cite{Wang18}, which is proven useful in text sentiment analysis. Table 5 shows the results of IAN$^+$ and Semi-IAN with and without the CRF attention layer. On both data sets, the accuracies of Semi-IAN with CRF attention mechanism have different degrees of reduction. Although the IAN$^+$ with CRF attention exhibits a certain improvement compared with the model without additional attention, its accuracy remains lower than that of our Semi-IAN. In summary, the CRF attention does not exhibit an outstanding performance in the experiment. The partial reason may lie in that the lengths of answers are usually short, whereas CRF attention is suitable for texts with moderate lengths. In our future work, we will investigate more attention mechanisms to solve this problem.
\begin{table}
\caption{Results of models with or without the CRF attention layer.}\label{tab5}\vspace{-0.11in}
\centering
\begin{tabular}{|c|c|c|c|}
\hline
Model & W/O CRF attention & rQAData1  & rQAData2 \\
\hline
IAN$^+$  & W & 0.722 & 0.548\\
IAN$^+$  & O & 0.720 & 0.578\\
Semi-IAN & W & 0.716 & 0.583\\
Semi-IAN & O & 0.735 & 0.585\\
\hline
\end{tabular}
\vspace{-0.11in}
\end{table}

%
%
%

\section{Conclusion}
We have investigated a new QA approach called rQA, in which machine is the questioner and human is the answerer. Human answer understanding is the key problem in rQA and is transformed into a classification problem in the present study. Two most common question types, namely, T/F and MC, are considered. Conventional text classification techniques are used to solve the answer understanding (or answer classification) in rQA. To elaborate the semantic relationships between questions and answers, IAN$^+$ is leveraged and an improved model called Semi-IAN is proposed. Semi-IAN considers the questions as the background and applies it into the deep feature representation for answers. Furthermore, two benchmark data sets are carefully constructed and made public. The experimental results indicate the initial success of the proposed models. Semi-IAN outperforms IAN$^+$ and methods based on conventional text classification techniques.
As rQA and its answer understanding are initially explored, there remains a number of challenges. Our future work will design more effective networks and introduce more domain knowledge.

\noindent \textbf{Acknowledgments.} This work is partially supported by NSFC $($61673377 and 61732011$)$, and Tianjin AI Funding $($17ZXRGGX00150$)$.

\end{document}